\documentclass[10pt,twocolumn,letterpaper]{article}

\usepackage{cvpr}      
\usepackage{float}
\usepackage{hyperref}
\usepackage{xcolor}
\usepackage[most]{tcolorbox}

\usepackage{booktabs}
\usepackage{amsmath}
\usepackage{graphicx}

\definecolor{cvprblue}{rgb}{0.21,0.49,0.74}
\def\paperID{*****} 
\def\confName{CVPR}
\def\confYear{2026}

\title{Self-Questioning Vision-Language Models: \\
Reinforcement Learning for Compositional Visual Reasoning}

\definecolor{codepurple}{HTML}{ADAA92}
\definecolor{softpink}{HTML}{CB8699}

\definecolor{refgreen}{HTML}{698D5B}

\hypersetup{
    colorlinks=true,
    linkcolor=black,      
    citecolor=softpink,      
    urlcolor=codepurple   
}
\author{
Saraswathy Amjith\\
Massachusetts Institute of Technology\\
\texttt{swathy@mit.edu}\\[0.75em]
{\large\textcolor{codepurple}{\textbf{Code: }}
\href{https://github.com/saraswathyamjith/sq-vlms}
{\textcolor{softpink}{\texttt{github.com/saraswathyamjith/sq-vlms}}}}
}

\begin{document}
\maketitle
\begin{abstract}
Vision-Language Models (VLMs) are AI systems that process both images and text, yet they often struggle with compositional visual reasoning questions that require chaining multiple steps together, such as identifying objects, counting them, and comparing the results. Existing approaches improve this reasoning by training models on human-written step-by-step explanations, but creating these annotations is expensive and difficult to scale. We propose a self-questioning framework that trains a VLM to break visual questions into smaller sub-questions and answer each one before producing a final response, using a reinforcement learning algorithm called Group Relative Policy Optimization (GRPO). The model is never shown examples of how to decompose questions, it discovers this behavior on its own, guided by a reward signal that scores whether the output contains sub-questions and whether the final answer is correct. We apply this framework to a 3-billion-parameter model, training on both synthetic scenes of geometric shapes (CLEVR) and real-world photographs (A-OKVQA).  On A-OKVQA, both self-questioning and standard reinforcement learning substantially improve accuracy over the untrained model (52.2\% and 51.6\% vs.\ 46.8\%). We introduce the first self-questioning VLM by rewarding not only the final answer like standard RL but additionally for generating intermediate sub-questions, enabling it to discover compositional decomposition strategies. These results suggest that teaching AI systems to ask themselves intermediate questions is a promising strategy for complex visual reasoning, particularly when the difficulty of a question warrants explicit step-by-step decomposition. 

    \end{abstract}

\section{Introduction}
\label{sec:intro}
A vision-language model uses transformers to encode images and text into a shared embedding space so that visual and linguistic information can be aligned, compared, or translated between modalities,  allowing a model to look at a photograph and
answer visual questions about it.  Recent advances in this
area include models such as GPT-4V~\cite{openai2023gpt4v},
LLaVA~\cite{liu2023llava}, and the Qwen-VL
family~\cite{bai2025qwen25vl}  which can now caption images, answer
open-ended questions about photographs, and follow complex multimodal
instructions~\cite{li2023seedbench}.  Recently, VLMS have been deployment in increasing complex real-world applications.

Yet much of this success involves questions that can be answered
through direct pattern recognition, identifying an object, reading
a label, or describing a scene.  A qualitatively harder class of
problems, compositional visual reasoning, remains a challenge. Compositional
visual reasoning is crucial to real-world applications such as autonomous-self driving or medical diagnosis. Consider the question, 'Is the pedestrian closer to the stop sign or the crosswalk?' about a street scene. A correct response requires that the model first locate the pedestrian, then identify the stop sign and crosswalk, estimate the distance to each, and finally compare the two. A medical imaging system asked 'Has the tumor grown since the last scan?' must locate the tumor in both images, measure its size in each, and compare the two.

State-of-the-art VLMs fail on compositional questions because they
produce answers in a single forward pass, i.e., the model processes the image and question once and immediately outputs an answer, without
any explicit intermediate reasoning.

Every existing approach to step-by-step visual reasoning depends on human-provided reasoning structure, limiting scalability.  These
strategies fall into three categories.  First, chain-of-thought
(CoT) prompting~\cite{wei2022cot, kojima2022zeroshotcot} and its
multimodal extensions~\cite{zhang2023multimodalcot, zheng2023ddcot}
supply the model with a handful of worked examples that demonstrate how to reason step by step, and then ask it
to follow the same pattern on new questions. While effective, these
methods are dependent on which examples are chosen, a poor selection can
significantly degrade performance, and the model does not internalize a
general reasoning strategy since the model weights are unchanged, it only imitates the specific patterns
shown in the prompt. Second,
supervised fine-tuning on human-written
reasoning ~\cite{lu2022scienceqa, chen2024m3cot}  updates the
model's weights by training it on datasets where humans have written out
each reasoning step.  This requires expensive, step-by-step annotations for every training
example, a bottleneck that scales poorly to large datasets.
Third, visual programming
approaches~\cite{gupta2023visprog, suris2023vipergpt} translate each
question into a short computer program that calls specialized vision
tools (for example, an object detector or a counter) in sequence.
These methods achieve strong compositional accuracy, but rely on developed
tool libraries, making them
lose effectiveness when faced with novel question types and generalization.

Reinforcement learning (RL) offers a different training strategy that can improve reasoning without explicit human-created reasoning traces. Unlike supervised approaches, which teach a model by showing it the ``right'' answer to imitate, RL lets the model generate multiple candidate responses to the same question, scores each response with a numerical reward (for example, $1$ for a correct final answer
and $-1$ for an incorrect one), and then adjusts the model's internal
weights so that higher-scoring responses become more likely in the
future while lower-scoring ones become less likely. In the text-only domain,
DeepSeek-R1~\cite{guo2025deepseekr1} demonstrated that a specific RL algorithm called Group Relative Policy Optimization
(GRPO)~\cite{shao2024deepseekmath} can train large language models
(LLMs) to generate intermediate reasoning steps without any
human-annotated reasoning traces, simply by rewarding correct final
answers produced in a structured format.   Intuitively, GRPO works by sampling a group of candidate responses for each prompt, scoring them against a reward signal, and then updating the model to favor higher-scoring responses relative to the group, eliminating the need for a separate critic or value model. 

We introduce a
self-questioning framework in which a VLM is trained via GRPO
to break visual questions into a sequence of sub-questions, answer each
sub-question by attending to the image, and then synthesize a final
response.  We require the model to produce output in a
structured format, alternating \texttt{<sub\_q>} (sub-question) and
\texttt{<sub\_a>} (sub-answer) tags, but we never show it example
decompositions during training.  Instead, we define a reward function
 that grants a score of
$1.0$ if and only if the model (a)~produces output in the 
sub-question format and (b)~arrives at the correct final answer, or a score of $-1.0$ otherwise.

We evaluate whether self-questioning emerges and transfers by training
on two benchmarks and testing across domains.  Our base model is
Qwen2.5-VL-3B-Instruct~\cite{bai2025qwen25vl}, a 3-billion-parameter
VLM that accepts both image and text inputs.  We chose this model
because its compact size makes RL training feasible on limited hardware,
while still having adequate existing capabilties. 
We train and evaluate on two benchmarks: CLEVR~\cite{johnson2017clevr}, a synthetic dataset of geometric scenes, and A-OKVQA~\cite{schwenk2022aokvqa}, a real-world visual question answering dataset (details in Section~\ref{sec:implementation}).

We investigate three questions: whether GRPO-based reinforcement learning improves visual question answering accuracy over the base model, whether the self-questioning format provides additional benefit beyond standard RL training, and whether the learned decomposition strategy transfers from synthetic to real-world visual domains.
Our ablations found that RL  is the primary reason for improvement. RL raises A-OKVQA accuracy from 46.8\% to 51.6\%, with self-questioning providing only an additional gain of +0.6\%, for an overall 52.2\%. A model trained with self-questioning on a dataset of synthetic images is able to transfer this skill when tested on real-world photographs (+2.6\% over baseline). The self-questioning format also introduces a cost on simpler tasks, where the overhead of generating sub-questions can decrease accuracy. These results suggest that the value of self-questioning is contingent on adequate question complexity, and that RL training itself accounts for a large portion of the observed gains. \\
\noindent\textbf{Contributions:} 
\begin{itemize}
    \item We trains a VLM to deconstruct compositional visual questions into sub-question--answer pairs using reinforcement learning, without any reasoning demonstrations.
    \item We conduct  experiments comparing self-questioning against direct RL on both synthetic (CLEVR) and real-world (A-OKVQA) benchmarks, identifying both the benefits and disadvantages of sub-questions.
\end{itemize}

\section{Related Work}
\label{sec:related}
Our work builds upon work in vision-language modeling, compositional visual reasoning, and reinforcement learning for language models.
Although modern vision-language models perform strongly on image understanding and open-ended visual question answering, they remain unreliable when asked to carry out multi-step reasoning over visual inputs. Step-by-step reasoning methods address this by introducing intermediate computations, but existing approaches depend on human-authored data to fine-tune on typically through SFT. 
Reinforcement learning samples multiple model responses and scores them using a reward function, and updates the model's weights to increase the likelihood of high-reward responses and decrease low-reward responses. Unlike supervised fine-tuning, RL requires no demonstrations of ideal reasoning, only an indication of whether the final answer is correct.

\paragraph{Vision-Language Models}
Large-scale VLMs such as Qwen2.5-VL-72B and GPT-4V exhibit some capabiltiies in compositional reasoning, answering questions that require combining multiple visual attributes, without task-specific training. However, this ability reduces significantly at smaller scales such as Qwen2.5-VL-3B-Instruct. We hypothesize that the model has sufficient capability in perception but an inability to decompose complex questions into sub-problems. To test this, we used reinforcement learning to train Qwen2.5-VL-3B-Instruct to generate sub-questions, enabling it to solve problems it previously answered incorrectly.

\paragraph{Multi-Step Reasoning in Vision-Language Models}
Several lines of work have attempted to improve VLM reasoning by introducing intermediate steps between question and answer. Chain-of-thought (CoT) prompting~\cite{wei2022cot} showed that language models produce more accurate answers when prompted to ``think step by step,'', but updated no model weights so it relies on base model capabilities.  Multimodal-CoT~\cite{zhang2023multimodalcot}, fine-tunes models on paired rationale generation and answer inference. DDCoT~\cite{zheng2023ddcot}, uses prompting to separate the reasoning phase from the answering phase. Self-taught reasoning (STaR)~\cite{zelikman2022star}, trains models to generate and filter their own reasoning traces through an iterative process, reducing dependence on human annotations but still requiring an initial set of correct rationales. All of these approaches are limited by their dependence on human-written reasoning demonstrations to guide the model's multi-step reasoning. In our framework, the model discovers how to pose sub- questions through reinforcement learning, with no reasoning demonstrations provided at all during training.

\paragraph{Reinforcement Learning for Reasoning}
DeepSeek-   R1~\cite{guo2025deepseekr1} demonstrated this in the text-only setting, showing that large language models develop chain-of-thought reasoning when trained with Group Relative Policy Optimization (GRPO)~\cite{shao2024deepseekmath} and rewarded only for correct final answers in a structured format. To our knowledge, our work is the first to apply GRPO to a vision-language model for compositional visual reasoning. 

No prior work has explicitly rewarded a model for decomposing visual questions into sub-questions, existing methods reward only the final answer or train on human-provided reasoning traces. Our framework makes decomposition itself part of the training objective by requiring sub-question–answer pairs for a positive reward.
\section{Method}
\label{sec:method}
\begin{figure}[H]
    \centering
    \includegraphics[width=.70\linewidth]{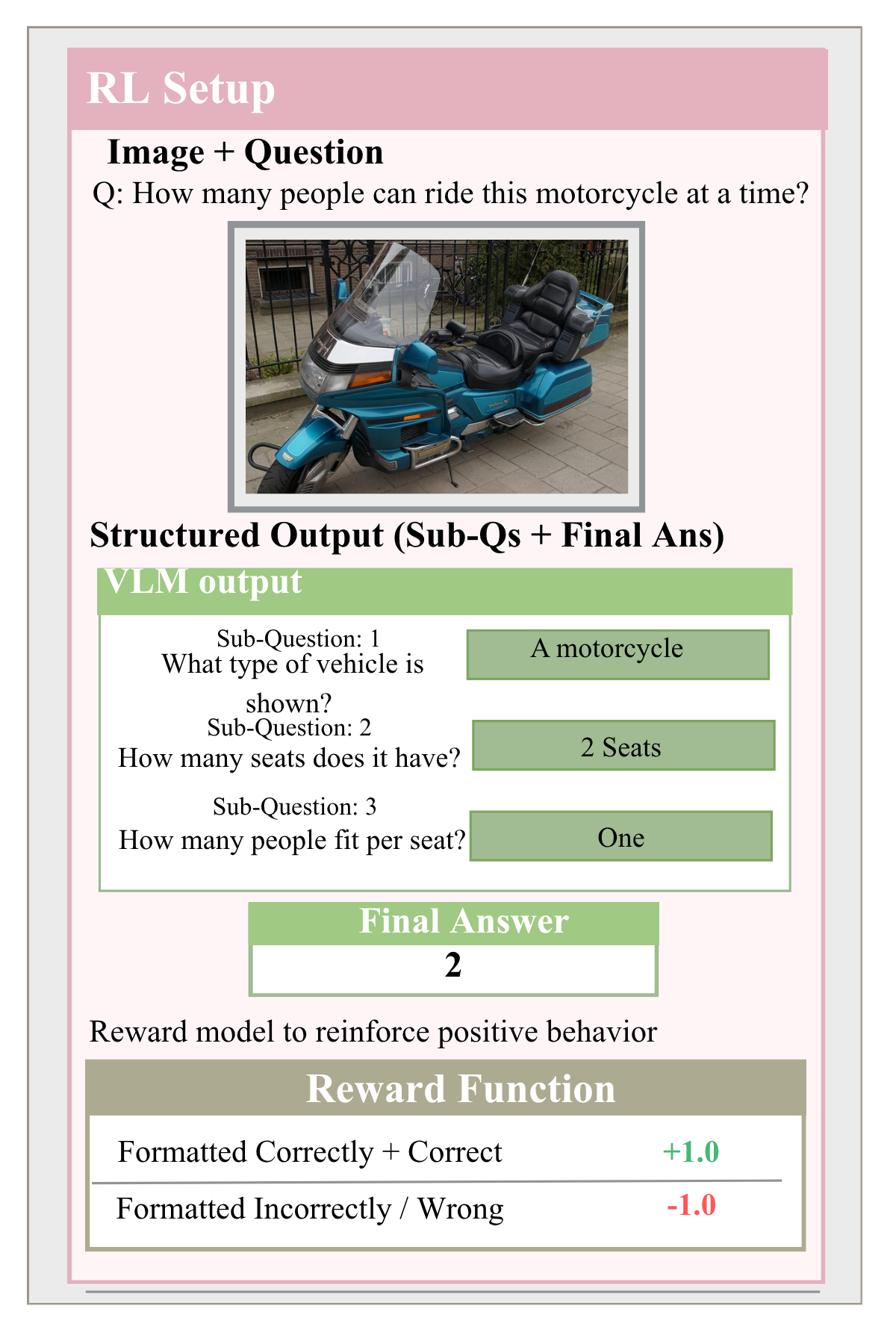}
    \caption{Overview of the self-questioning framework.  Given an image and
question, the VLM generates sub-questions, answers each by attending to
the image, and produces a final answer.  The reward function requires
both correct format and correct answer; GRPO uses group-normalized
advantages to update the policy.}
    \label{fig:method_overview}
\end{figure}

We investigate whether a VLM can learn to decompose compositional visual questions into sub-questions entirely through reinforcement learning, without any human-authored reasoning demonstrations. Our approach addresses the annotation bottleneck identified in Section~\ref{sec:intro} by replacing human-written reasoning traces with a reward signal that incentivizes both sub-question generation and correct answers, allowing the decomposition behavior to emerge from training. Figure~\ref{fig:method_overview} provides an
overview of the RL setup.

\subsection{Problem Formulation}
\label{sec:problem}

Given an image $I$ and a question $q$, a VLM generates a textual response
$y = f_\theta(I, q)$.  In standard Visual-Question-Answering, the model produces a direct
answer in one step.  We instead train the model to generate a structured response
consisting of sub-question--answer pairs followed by a final answer:
\begin{align}
y &= [\text{SQ}_1, \text{SA}_1, \ldots, \text{SQ}_k, \text{SA}_k, \text{FA}]
\end{align}
where $\text{SQ}_i$ and $\text{SA}_i$ denote the $i$-th sub-question and
its answer, $k$ is the number of sub-questions (determined by the model,
not by us), and $\text{FA}$ is the final answer.  

\subsection{Self-Questioning Prompt}
\label{sec:prompt}
We use a system prompt that instructs the model to generate sub-questions
before answering.  

\definecolor{promptgreen}{HTML}{A1BB93}
\definecolor{promptpink}{HTML}{E4B1BF}
\definecolor{promptgold}{HTML}{C4BF7E}

\begin{tcolorbox}[
    enhanced,
    breakable,
    colback=promptpink!25,
    colframe=promptgreen,
    boxrule=0.8pt,
    arc=2mm,
    left=6pt,
    right=6pt,
    top=6pt,
    bottom=6pt,
    borderline west={2pt}{0pt}{promptgold}
]
\tiny\ttfamily
You are given an image and a question. Before answering, generate a series
of sub-questions to help you analyze the image carefully. For each
sub-question, look at the image and provide an answer. Then give your
final answer.

\vspace{0.5em}

\textcolor{promptgold}{Format:}\\
Sub-question 1: [question] \\
Answer 1: [answer] \\
Sub-question 2: [question] \\
Answer 2: [answer] \\
\ldots \\
Final Answer: [answer]
\end{tcolorbox}
We deliberately omit examples from the prompt because including
them would bias the model toward imitating specific decomposition
patterns rather than discovering its own. 

\subsection{Reward Function}
\label{sec:reward}
Our reward function enforces format compliance and
answer correctness.  Given a model completion $y$ and
ground-truth answer $a^*$:
\begin{equation}
R(y, a^*) =
\begin{cases}
1.0 & \text{if } \text{Format}(y) \land \text{Correct}(y, a^*)  \\
-1.0 & \text{otherwise}
\end{cases}
\label{eq:reward}
\end{equation}
where $\text{Format}(y)$ checks for at least one
``Sub-question / Answer'' pair, $|\text{FA}|$ is the word count of the
final answer, and $\text{Correct}(y, a^*)$ uses containment
matching, the normalized ground truth must appear as a substring of the
normalized prediction (case-insensitive).
Unlike prior RL approaches that reward only final-answer correctness, our reward function makes the decomposition structure itself a training objective.

\subsection{Training with GRPO}
\label{sec:grpo}

We use Group Relative Policy Optimization
(GRPO)~\cite{shao2024deepseekmath} as our RL algorithm.  We chose GRPO
over the more common Proximal Policy Optimization
(PPO)~\cite{schulman2017ppo} because GRPO eliminates the need for a
separate value network (a second neural network that estimates how good
each state is), reducing memory requirements by roughly half, a
critical advantage when training on limited GPU hardware.  Instead of
learning a value baseline, GRPO normalizes rewards within groups of
completions generated for the same prompt.

For each training prompt $(I, q)$, GRPO samples $G$
completions $\{y_1, \ldots, y_G\}$ from the current model and computes
group-normalized advantages:
\begin{equation}
\hat{A}_i = \frac{R(y_i, a^*) - \mu_g}{\sigma_g}
\label{eq:advantage}
\end{equation}
where $\mu_g$ and $\sigma_g$ are the mean and standard deviation of
rewards within the group.   The policy is then updated using a  objective with a KL (Kullback-Leibler) divergence penalty against the reference
policy $\pi_{\text{ref}}$:
\begin{equation}
\begin{split}
\mathcal{L}(\theta) = -\mathbb{E}\Big[
  &\min\!\big(r_t \hat{A}_t,\;
  \text{clip}(r_t, 1{-}\epsilon, 1{+}\epsilon)\hat{A}_t\big) \\
  &- \beta\, D_{\text{KL}}(\pi_\theta \| \pi_{\text{ref}})
\Big]
\end{split}
\end{equation}
where $r_t = \pi_\theta(y_t) / \pi_{\text{ref}}(y_t)$ is the importance
sampling ratio.  
The KL divergence measures how far the updated policy has drifted from the reference policy. The $\beta$-weighted penalty prevents the model from changing too heavily during training, without it, the model might improve at answering questions in the sub-question format but lose its underlying ability to understand images or produce coherent language, a phenomenon known as catastrophic forgetting.

Our training procedure follows DeepSeek-R1's application of GRPO to elicit structured reasoning, but diverges in two ways: (1) we operate in the multimodal setting, requiring the model to attend to image inputs during sub-question generation, whereas DeepSeek-R1 trains text-only LLMs; (2) our reward function  requires sub-question–answer pair formatting rather than rewarding free-form chain-of-thought, as we hypothesize that enforcing decomposition into sub-questions improves compositional reasoning more effectively.

\subsection{Baseline: Direct RLVR}
\label{sec:baseline}
To isolate the specific effect of self-questioning from the general
effect of RL training, we train a baseline model
using the same GRPO procedure but with a simplified prompt that requests
only a direct answer and a reward function that
checks answer correctness and conciseness without any format requirement.
This baseline is essential because it represents standard reinforcement
learning from verifiable rewards (RLVR) applied to visual question answering.  By using
identical hyperparameters and training data for both models, any
accuracy difference between them can be attributed specifically to the
self-questioning format rather than to RL training in general.

\subsection{Implementation Details}
\label{sec:implementation}

\paragraph{Base model.} 

We use Qwen2.5-VL-3B-Instruct~\cite{bai2025qwen25vl},
a 3-billion-parameter VLM.
Its compact size makes RL training feasible on academic-scale hardware
while still being
large enough to exhibit meaningful visual reasoning capabilities.

\paragraph{Datasets} 

We train on two datasets:
(1)~a 10{,}000-example subset of CLEVR~\cite{johnson2017clevr}, a
synthetic visual reasoning benchmark with compositional questions about
computer-generated scenes of geometric objects; and
(2)~a 10{,}000-example subset of A-OKVQA~\cite{schwenk2022aokvqa}, a
challenging real-world visual question answering (VQA) dataset requiring outside knowledge and
complex reasoning about natural photographs.  CLEVR's simpler questions make it an ideal starting point for studying emergent visual reasoning, while A-OKVQA tests whether the framework generalizes to the greater complexity of natural images, which demand external knowledge beyond what is visually present. We selected CLEVR in part because of efficiency, its synthetic, simple scenes allow rapid training. We selected A-OKVQA because of its universality, its diverse real-world photographs and open-ended questions test whether models can generalize across the broad range of visual and knowledge domains that practical deployment demands.

\begin{figure}[H]
    \centering
    \includegraphics[width=.70\linewidth]{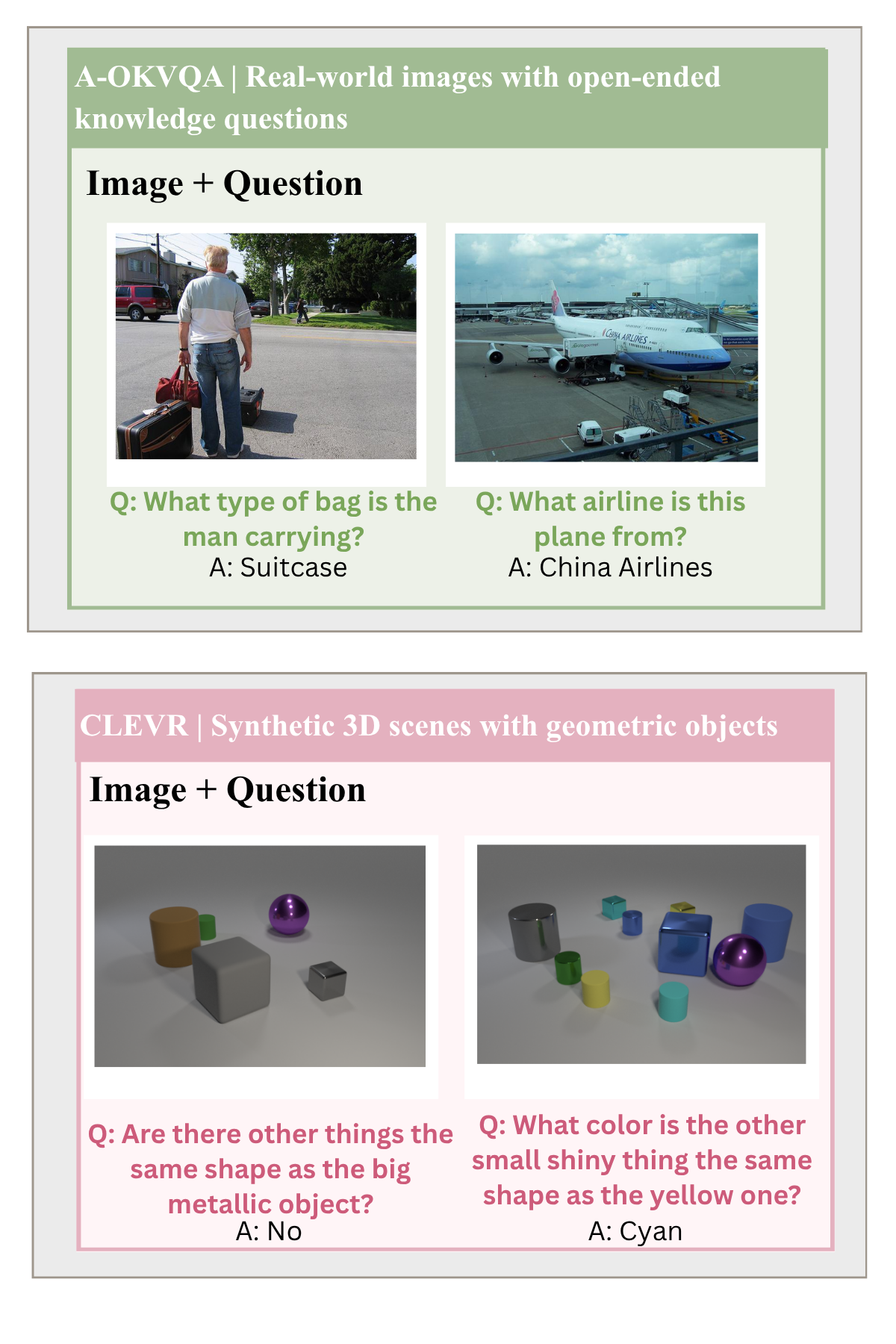}\caption{Examples from the two evaluation benchmarks.\\ \textbf{Top:} A-OKVQA, which pairs real-world images with open-ended questions requiring external knowledge. \\\textbf{Bottom:} CLEVR, which poses compositional reasoning questions over synthetic 3D scenes with geometric objects.}
\label{fig:benchmark-examples}
\end{figure}

\begin{figure}[H]
    \centering
    \includegraphics[width=.70\linewidth]{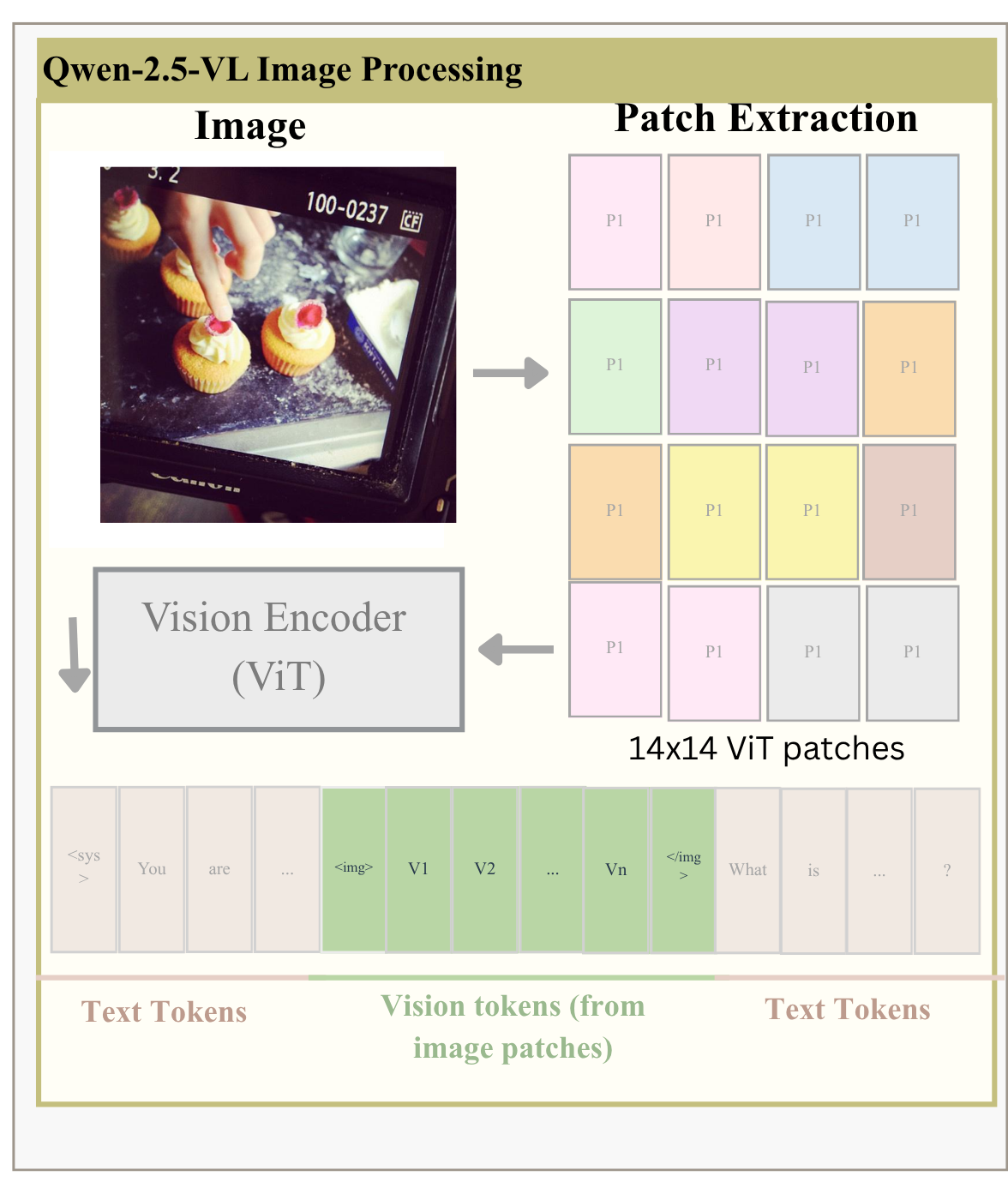}
    \caption{Overview of image processing in Qwen2.5-VL. The input image is divided into 14$\times$14 patches, which are encoded by a Vision Transformer (ViT) into a sequence of vision tokens. These vision tokens are interleaved with text tokens in the model's input sequence, delimited by special \texttt{<img>} tokens.}
\label{fig:qwen-vl-processing}
\end{figure}

\paragraph{Parameter-efficient fine-tuning.} Rather than updating all
3 billion parameters during training, we apply Low-Rank Adaptation
(LoRA)~\cite{hu2022lora}. In standard fine-tuning, every weight in the
model is adjusted, which requires storing a full copy of all parameter
gradients in memory. LoRA instead freezes the original weights and
learns a small pair of matrices for each targeted layer that together
approximate the weight change needed for the new task. Because these
matrix pairs are much smaller than the original weight matrices, LoRA
updates less than 1\% of the model's parameters while achieving
performance comparable to full fine-tuning~\cite{hu2022lora}. We use
rank $r=16$, $\alpha=32$, and dropout $0.05$, targeting the query and
value projection matrices (\texttt{q\_proj}, \texttt{v\_proj}). This
fits within the memory capacity of our GPUs (48\,GB each), which full
fine-tuning would exceed.
\section{Results}
\label{sec:results}
We designed experiments to answer three
questions: (1)~Does Reinforcement-Learning training improve Visual Question Answering accuracy? (2)~Does
self-questioning provide additional benefit beyond just Reinforcement-Learning? (3)~Does
the learned reasoning behavior transfer across visual domains?

\subsection{Experiments} We compare two prompt formats: self-questioning (SQ), in which the model must produce sub-question–answer pairs before a final answer, and direct, in which the model produces only a final answer with no intermediate decomposition. 
We evaluate five models
(See Our Appendix Table~\ref{tab:hyperparams} for Hyperparameters) for these comparisons:
\begin{itemize}
    \item \textbf{SQ+GRPO (CLEVR)}: trained w/ self-questioning prompt on
    CLEVR with $G{=}8$ for ${\sim}$7{,}000 steps.
    \item \textbf{Direct+GRPO (CLEVR)}: trained w/ direct-answer prompton
    CLEVR with $G{=}8$ for ${\sim}$7{,}000 steps.
    \item \textbf{SQ+GRPO (A-OKVQA)}: trained w/ self-questioning prompt on
    A-OKVQA with $G{=}8$ for ${\sim}$7{,}000 steps.
    \item \textbf{Direct+GRPO (A-OKVQA)}: trained w/ direct-answer prompt on
    A-OKVQA with $G{=}8$ for ${\sim}$7{,}000 steps.
    \item \textbf{Base}: the unmodified Qwen2.5-VL-3B-Instruct model,
    with no RL training applied.
\end{itemize}
On each dataset, the SQ+GRPO and Direct+GRPO models use identical
hyperparameters (Table~\ref{tab:hyperparams}) and training data, differing only in prompt format and their 
reward function, ensuring
any performance difference between the two models can be attributed
to the self-questioning structure itself, rather than to differences in
training data, compute, or optimization. To test whether the learned decomposition strategy is domain-general, we evaluate the CLEVR-trained SQ model on A-OKVQA and vice-versa. To distinguish whether accuracy gains come from improved internal reasoning or simply from the prompt format, we evaluate the direct-RL model with a self-questioning prompt.

\subsection{Evaluations}
\label{sec:eval_protocol}
We evaluate using 500 examples from the validation set of each benchmark.  We report accuracy (the fraction of correctly answered questions) in direct-answer mode, where the model generates a free-form response which is then checked for containing the answer as a sub-string.

\subsection{A-OKVQA: Main Comparison}
\label{sec:aokvqa_results}

\begin{table}[H]
\centering
\small
\caption{A-OKVQA validation accuracy (500 examples).  SQ+GRPO and
Direct+GRPO are both trained on A-OKVQA for ${\sim}$7{,}000 matched steps.
Both RL methods improve over the base model, with SQ providing a
marginal advantage.}
\label{tab:aokvqa_results}
\begin{tabular}{llcc}
\toprule
Model & Train Data & Eval Prompt & Accuracy \\
\midrule
Base & --- & direct & 46.8 \\
\midrule
Direct+GRPO & A-OKVQA & direct & 51.6 \\
SQ+GRPO & A-OKVQA & sq & \textbf{52.2} \\
SQ+GRPO & A-OKVQA & direct & 47.8 \\
\midrule
Direct+GRPO & A-OKVQA & direct & 47.0 \\
SQ+GRPO & CLEVR & sq & \textbf{49.4} \\
SQ+GRPO & CLEVR & direct & 47.4 \\
\bottomrule
\end{tabular}
\end{table}

\begin{table}[H]
\centering
\small
\caption{CLEVR validation accuracy (500 examples).  All models
near ceiling on this benchmark.  Self-questioning
introduces a format tax: the A-OKVQA-trained SQ model drops to 81.0\%
with the SQ prompt but recovers to 97.6\% with a direct prompt,
confirming the model weights are intact.}
\label{tab:clevr_results}
\begin{tabular}{llcc}
\toprule
Model & Train Data & Eval Prompt & Accuracy \\
\midrule
Base & --- & direct & 97.4 \\
\midrule
Direct+GRPO & CLEVR & direct & 98.4 \\
SQ+GRPO & CLEVR & direct & \textbf{98.6} \\
SQ+GRPO & CLEVR & sq & 97.2 \\
\midrule
Direct+GRPO & A-OKVQA & direct & \textbf{98.6} \\
SQ+GRPO & A-OKVQA & direct & 97.6 \\
SQ+GRPO & A-OKVQA & sq & 81.0 \\
\bottomrule
\end{tabular}
\end{table}

\begin{figure}[H]
\centering
\includegraphics[width=\columnwidth]{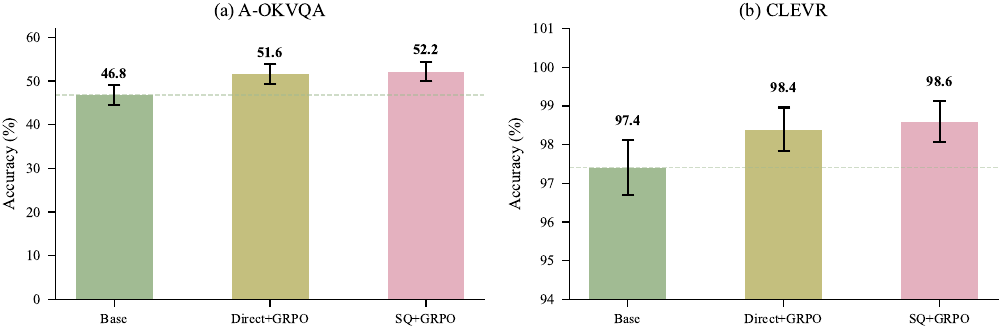}
\caption{Accuracy comparison across A-OKVQA and CLEVR.  Both RL methods
improve over the base model on A-OKVQA, but the SQ format introduces a
format tax on CLEVR that disappears with direct prompting, confirming
the tax is a property of the prompt, not the model weights.}
\label{fig:accuracy_bars}
\end{figure}

\begin{figure}[H]
\includegraphics[width=\columnwidth]{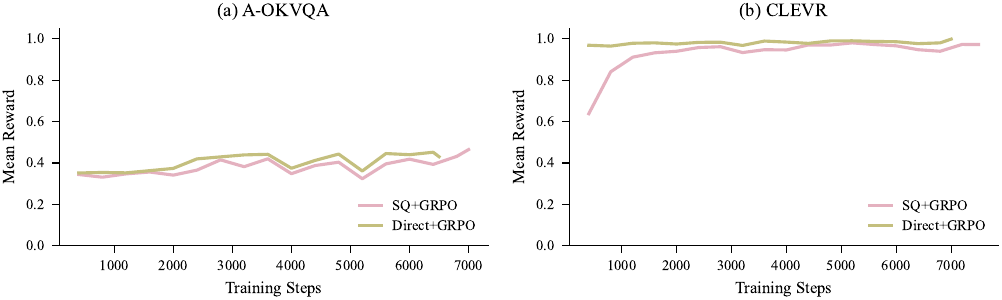}
\caption{Training reward curves ($G{=}8$, 200-step rolling average).
\textbf{(a)}~On A-OKVQA, both methods gradually improve, with SQ+GRPO
and Direct+GRPO reaching similar reward levels by 7{,}000 steps.
\textbf{(b)}~On CLEVR, Direct+GRPO starts near 0.97 while SQ+GRPO
climbs from 0.64 as the model learns the structured format.}
\label{fig:training_curves}
\end{figure}

\paragraph{RL training improves accuracy.}
Both GRPO-trained models substantially outperform the base model
(Figure~\ref{fig:accuracy_bars}, Table~\ref{tab:aokvqa_results} Table~\ref{tab:clevr_results}): SQ+GRPO achieves \textbf{52.2\%}
(+5.4\%) and Direct+GRPO achieves \textbf{51.6\%} (+4.8\%), compared
to the base model's 46.8\%.
GRPO reinforces correct responses while suppressing incorrect ones, concentrating the model's output distribution around high-confidence answers over thousands of training steps. Recent work~\cite{yue2025limit-of-rlvr} has shown that RLVR improves pass@1 primarily through biasing the model towards reliable, correct answers at the cost of reduced output diversity.  Both CLEVR and many A-OKVQA questions can be answered correctly through well-calibrated single-step responses, without requiring explicit decomposition. This likely explains why RL alone accounts for the majority of our observed improvement.

\paragraph{Self-questioning provides marginal benefit.}
At matched training steps, SQ+GRPO (52.2\%) slightly outperforms
Direct+GRPO (51.6\%), a difference of only 0.6 percentage points.
This suggests that the accuracy gain comes primarily from
the RL training signal rather than from the structured sub-question
decomposition.  One likely explanation is that A-OKVQA questions,
while challenging, often do not require deep multi-step decomposition:
many can be answered through strong visual recognition and world
knowledge, which RL training improves regardless of output format.
The training reward curves
(Figure~\ref{fig:training_curves}) illustrate how on CLEVR, the
Direct+GRPO model starts with a reward near 0.97, leaving almost no
room for self-questioning to provide additional benefit. On A-OKVQA,
both methods start lower and converge to similar reward levels by
7{,}000 steps, 
\paragraph{Prompt ablation.}
When the A-OKVQA-trained SQ model is evaluated with a direct prompt instead of the SQ prompt, accuracy drops to 47.8\%, nearly back to the base model level of 46.8\%. This is expected. During training, every prompt included the sub-question format, so RL reinforced weight updates conditioned on that specific prompt structure. The model's weights have been optimized to produce high-reward responses given the SQ prompt, when a direct prompt is provided instead, those weight updates are not elicited, and accuracy reverts to near-baseline levels.

\paragraph{Cross-dataset transfer.}
The CLEVR-trained SQ model achieves 49.4\% on A-OKVQA with the SQ
prompt (+2.6\% over base), despite never seeing real-world images
during training. In contrast, the CLEVR-trained Direct+GRPO model
achieves only 47.0\% on A-OKVQA (+0.2\% over base), suggesting that
standard RL learns improvements largely specific to the training
domain. CLEVR contains only synthetic scenes of geometric shapes, yet
the self-questioning strategy learned on these simple images transfers
effectively to complex natural photographs. This gap indicates that
what transfers is the decomposition strategy itself, a domain-general
skill, rather than the answer-level reinforcement that standard RL
provides.

\paragraph{Capability loss on simpler tasks.}
Table~\ref{tab:clevr_results} presents CLEVR validation results.
Self-questioning introduces what we call a format tax, an
accuracy cost from the overhead of generating sub-questions on problems
simple enough to answer directly. The A-OKVQA-trained SQ model achieves
only \textbf{81.0\%} on CLEVR with the SQ prompt, a 16.4\% drop from
the base model's 97.4\%. However, when the same model is evaluated
with a direct prompt, accuracy recovers to 97.6\%, confirming that
the model's visual capabilities remain, the loss comes entirely
from forcing decomposition on questions like ``What shape is the matte object?'' that the model can already answer in a single step, but additional questions can produce noise.
The CLEVR-trained SQ model pays a much smaller format tax: 97.2\% with
the SQ prompt vs.\ 98.6\% with a direct prompt (only 1.4\%). Training
on in-domain data teaches the model to generate  sub-questions
that add little overhead, suggesting the tax is reducible with
in-domain training.

\subsection{Qualitative Analysis}
\label{sec:qualitative}
To understand what the model actually learns to do, we examine
individual model outputs.
The SQ-trained model generates sub-questions for every response (100\%
format compliance), confirming that the format constraint is fully
learned.  On A-OKVQA, the sub-questions sometimes identify relevant
aspects of the scene.  For example:
\begin{quote}
\small\ttfamily
Sub-question 1: What are the people in the image doing? \\
Answer 1: They are skiing down a mountain. \\
Sub-question 2: Where are they located? \\
Answer 2: They are on the slope. \\
Final Answer: Mountain peak.
\end{quote}
Here, the sub-questions decompose into intermediate
observations (activity, location) that support the final
answer. However, on CLEVR the decompositions often restate the original question nearly verbatim. This likely reflects the simplicity of CLEVR questions, since the base model can already answer most of them correctly in a single step, so there is no reasoning gap for sub-questions to fill. 
\section{Discussion}
\label{sec:discussion}

\subsection{Key-Findings}
Our results reveal a more nuanced picture than the introduction's hypothesis suggested. We motivated this work by arguing that VLMs fail at compositional reasoning because they produce answers in a single forward pass without intermediate reasoning steps. However, standard RL training without any sub-question structure accounts for the majority of the improvement (+4.8\%), while self-questioning adds only +0.6\% on A-OKVQA. This suggests that much of the base model's failure on compositional questions is not due to a missing reasoning structure, but from a poorly calibrated output distribution, RL corrects this by reinforcing correct answers regardless of format.
That said, self-questioning shows clear value in specific settings. The cross-domain transfer result, where a model trained only on synthetic geometric shapes improves by +2.6\% on real-world photographs, suggests that the decomposition strategy itself generalizes, independent of visual domain knowledge. Self-questioning also provides the largest gains on question types that require the kind of multi-step reasoning our motivating examples illustrate: counting (+50\% over base) and material identification (+33\%), tasks where a single-step answer is  unreliable and intermediate observations help. Conversely, the format tax on simpler tasks like color recognition and spatial relations confirms that decomposition is counterproductive when the question can already be answered in one step.
These findings suggest that the value of self-questioning is contingent on question complexity. Future work may benefit from adaptive prompting strategies that apply self-questioning selectively based on estimated question difficulty, decomposing only when the overhead is likely to pay off.

\paragraph{Comparison with related approaches.} To our knowl- edge, no prior work has applied GRPO to a vision-language model at the 3B-parameter scale for compositional reasoning.
 Our +5.4\% gain on A-OKVQA is achieved without human-authored reasoning traces, unlike supervised methods such as Multimodal-CoT~\cite{zhang2023multimodalcot} that require paired rationale annotations. DeepSeek-R1~\cite{guo2025deepseekr1} showed GRPO elicits chain-of-thought in text-only LLMs, our results partially replicate this in the multimodal setting. Visual programming methods~\cite{ suris2023vipergpt} achieve strong compositional accuracy but depend on hand-designed tool libraries, whereas our framework requires no external tools and transfers across domains. However, our failure cases reveal a limitation of purely emergent reasoning: the model occasionally generates superficial sub-questions that paraphrase rather than decompose, suggesting that while RL can teach a model to adopt a reasoning format, it does not guarantee the quality of the reasoning within that format. Perhaps future work will benefit from understanding the balance between fully human-authored and fully emergent reasoning.

\subsection{Failure Cases}
\label{sec:failures}
We identify two recurring failure cases in SQ+GRPO outputs.
\paragraph{Superficial sub-questions.}
The model sometimes generates sub-questions that paraphrase the original
question rather than decomposing it.  For example, given ``What sport is
being played?'', the model may ask ``What activity are the people doing?''
and ``What game is this?'', i.e. restatements that do not extract new
visual information.  The sub-questions in this case are only adding redundant tokens.
\paragraph{Format-induced errors on simple questions.}
Self-questioning introduces errors when decomposition is unnecessary.
On CLEVR, the A-OKVQA-trained SQ model drops to 81.0\% with the SQ
prompt, as the model decomposes questions like ``What color is the large metallic
sphere?'' that it can already answer in a single step.  On A-OKVQA, self-questioning answers questions the baseline
couldn't in 14\% of cases (69/500) but fails on 8\% (42/500) that the
base model answers correctly.  The failures concentrate on simple
questions such as color recognition ($-$37.5\%), spatial relations
($-$13.3\%), and activity identification ($-$16.7\%), where any
intermediate step introduces opportunities for propagated errors. In many cases, the model identified the correct concept in a sub-answer but rephrased
it to a non-matching term in the final answer.  However, self-questioning helps when
multi-step reasoning is needed: counting (+50\% over base) and 
material identification (+33\%). 

\subsection{Limitations}
\paragraph{Model scale.}
We train a 3B-parameter model with LoRA.  Larger models or full
fine-tuning may show different results. Self-questioning could
provide greater benefit for small models with weaker base reasoning, because
those models have more room for improvement through structured
decomposition.  Conversely, very large models may already reason
compositionally without explicit structure. Future work involves testing
across model sizes (e.g., 7B, 14B, 72B variants in the Qwen-VL family)
with full fine-tuning to reveal where self-questioning provides the
greatest improvement.

\paragraph{Evaluation scope.}
We evaluate on two VQA benchmarks.  Tasks requiring deeper multi-step
reasoning  may
show larger benefits from structured decomposition, because the gap
between what the model can answer in one step versus multiple steps
is wider on harder tasks.

\paragraph{Transparency and reproducibility.}
We use a pre-trained open-weight model (Qwen2.5-VL-3B-Instruct) and
publicly available datasets.  All hyperparameters are reported in
Table~\ref{tab:hyperparams}, and our code is available to
support reproducibility.

\section{Conclusion}
\label{sec:conclusion}
We investigated whether a VLM could learn to pose  sub-questions without any human demonstrations of
multi-step reasoning. Self-questioning and standard RL training substantially improve over the base model, raising A-OKVQA accuracy from 46.8\% to 52.2\% and 51.6\% respectively. RL training accounts for the majority of this improvement, with the sub-question format adding only a marginal benefit (+0.6\%) and actually reducing accuracy on simpler tasks where decomposition creates overhead. However, self-questioning shows a clear advantage over standard RL in cross-domain transfer. the CLEVR-trained SQ model improves by +2.6\% on real-world A-OKVQA photographs, while the CLEVR-trained direct RL model gains only +0.2\%, suggesting that sub-question decomposition teaches a domain-general reasoning strategy that transfers more effectively than the answer-level reinforcement of standard RL.  These findings point toward a practical direction of adaptive systems
that apply self-questioning selectively, decomposing only when question
complexity warrants the overhead, an approach particularly relevant to safety-critical domains like autonomous driving and medical imaging where multi-step visual reasoning is essential.  Our framework demonstrates that the ability to decompose complex visual questions into sub-tasks can emerge from reward signals alone, offering a scalable alternative to the costly human annotations that current multi-step reasoning approaches require.

{
    \small
    \bibliographystyle{ieeenat_fullname}
    \bibliography{main}
}
\newpage
\appendix

\section{Training Hyperparameters}

\begin{table}[H]
\centering
\small
\caption{GRPO training hyperparameters, matched across all models and
datasets.}
\label{tab:hyperparams}
\begin{tabular}{lc}
\toprule
Hyperparameter & Value \\
\midrule
Base model & Qwen2.5-VL-3B-Instruct \\
LoRA rank / alpha & 16 / 32 \\
LoRA targets & \texttt{q\_proj}, \texttt{v\_proj} \\
Learning rate & $1 \times 10^{-5}$ \\
Batch size (effective) & 4 \\
$\beta  \& 0.01$ \\
Generations per prompt ($G$) & 8 \\
Max completion length & 256 tokens \\
Sampling temperature & 0.7 \\
Training steps & ${\sim}$7{,}000 \\
Training examples & 10{,}000 \\
Optimizer & AdamW \\
Precision & bfloat16 \\
\bottomrule
\end{tabular}
\end{table}

\end{document}